# A Constraint-Driven System for Contract Assembly

Aspassia Daskalopulu     Marek Sergot

**Department of Computing**
**Imperial College of Science, Technology and Medicine**
**180 Queen's Gate, London SW7 2BZ**
E-mail: {ad4,mjs}@doc.ic.ac.uk

**ABSTRACT**

We present an approach for modelling the structure and coarse content of legal documents with a view to providing automated support for the drafting of contracts and contract database retrieval. The approach is designed to be applicable where contract drafting is based on model-form contracts or on existing examples of a similar type. The main features of the approach are: (1) the representation addresses the structure and the interrelationships between the constituent parts of contracts, but not the text of the document itself; (2) the representation of documents is separated from the mechanisms that manipulate it; and (3) the drafting process is subject to a collection of explicitly stated constraints that govern the structure of the documents. We describe the representation of *document instances* and of *'generic documents'*, which are data structures used to drive the creation of new document instances, and we show extracts from a sample session to illustrate the features of a prototype system implemented in MacProlog.

## 1  INTRODUCTION

The system described in this paper forms part of a broader project to develop automated support tools for the drafting, management and administration of large, complex contracts. The term 'contract' is intended to be understood here in its common usage, referring sometimes to a (legally binding) agreement and sometimes to the document in which this agreement is recorded. Where it is important to distinguish between these meanings we shall use the terms 'agreement' and 'document' accordingly.

The contracts that we have been using as experimental material concern the purchase, supply and transportation of natural gas. In common with contracts in many engineering fields, these contracts are large: they cover details of pricing and payment, supply schedules, quality assurance, monitoring, maintenance of equipment, force majeure provisions, and so on; a typical document will run to 200 or 300 pages, not including detailed drawings and technical appendices. (Some previous projects on the automation of legal document drafting have looked towards areas such as Sale of Goods (see e.g. [ALDUS 1992]) for their applications, and have found limited opportunities for automation. We believe that domains concerned with the provision of engineering services of various kinds are likely to be more fruitful. Contracts of the size we have been dealing with are routinely encountered; we have anecdotal evidence that there are engineering projects where the associated contracts are an order of magnitude larger than the ones we have been examining, and where the drafting and administration of the contract accounts for a significant portion of the total cost of the project.)

Apart from their size, the contracts are also complex, in two senses: they consist of a number of separate but closely inter-related sub-agreements, and they are complex at the micro level, in the sense that some of the provisions contain a mass of complicated detail which is both difficult to follow and difficult to apply in specific cases.

By 'management of contracts' we refer to the general problem of storing, maintaining and retrieving large bodies of contracts and associated documents. A system for searching a document database for relevant contracts and parts of contracts would fall in this category. Under 'administration' we have in mind systems that perform more specific tasks: systems that advise on the effects of detailed provisions, or on the procedures to be followed in certain circumstances, and systems which monitor compliance of the contracting parties with detailed requirements of the agreement. We are aiming to provide drafting tools to support the design and drafting of contracts at the micro level—that is, the formulation of detailed provisions—and at the macro level where the problem can be thought of as deciding what components need to be included in a contract and in what form.

In this paper we focus on the drafting of contracts at the macro level, where the emphasis is on structure and overall coherence. Contracts are represented at a coarse level of detail. We make no attempt—in this paper—to provide a representation of what a contract actually *says* or *prescribes* (what it 'means'). The text of the documents, in appropriate fragments, are taken as atomic units of the representation and are not subjected to any further form of analysis.

In general terms, contract drafting is viewed here as a form of Computer-Aided Design, where the drafter uses basic blocks of text to construct a document in much the same way that a graphics designer uses basic geometric shapes to construct a picture or diagram. It may be more appropriate to call such a process contract 'assembly' rather than contract 'drafting' to emphasise the use of pre-constructed building blocks. A similar view has been expressed by [Fiedler 1985] and [Gordon 1989]. Lauritsen [1992] draws the analogy between legal document assembly and the configuration of a computer system. Lauritsen's article [1992] also provides a very useful survey of previous approaches to computerised legal document drafting generally.

The best known early system for legal document drafting is Sprowl's ABF processor (see e.g. [Sprowl 1980]) designed for the automated assembly of wills, tax returns, trusts and other standardised documents.

However, Sprowl's approach is procedural and comparatively low-level, since in effect he provides a special-purpose imperative programming language for writing programs which can generate certain types of documents (cf. [Gordon 1989]). Fragments of legal documents, which are subject to change, depending on specific data, are identified and the text is encoded with suitable procedures for filling in the values of these parameters.

Another well-known document assembly system is Scrivener, described by Lauritsen [1992] as "an expert system shell with query-the-user facilities [...] specifically adapted for text generation". Scrivener offers users more flexibility than the ABF processor and the representation of documents is not entirely procedural. However, as in the ABF processor, mechanisms to insert values for specific parameters and conditions for the inclusion of specific document components are incorporated in the text itself.

Our approach in contrast relies on an explicit representation of document structure. We separate the representation from the mechanisms used to create new instances, and we formulate *explicit* constraints which govern the structure of a given document type. Creation of a new document instance is the process of assembling suitably instantiated blocks of text, subject to compliance with these constraints.

We want to represent what each unit of the contract contributes and why it has been included in its chosen form. As a secondary, though still important, objective we want to support the storage and retrieval of large bodies of contracts.

## 2  DOCUMENT INSTANCES

We begin by describing the output of the drafting system, which is a database of document *instances*.

Each document instance records:

- the type of document and an identifier for the specific document instance;

- a term containing information about the contracting parties and the date on which the contract came into effect;

- optionally, other data values (depending on the document type); and

- a list of terms corresponding to the sub-units of the document.

Each of the sub-units is also represented as a term with similar structure: type, identifier, data values (if any), and further sub-divisions. A more detailed example of the representation is provided later in the paper.

The resulting representation is a tree-like, skeletal structure similar to those often used for hierarchically structured documents (cf. [Furuta 1989]). The tips of this structure correspond to actual fragments of text (or rather pointers into text files stored separately on disk), as illustrated below:

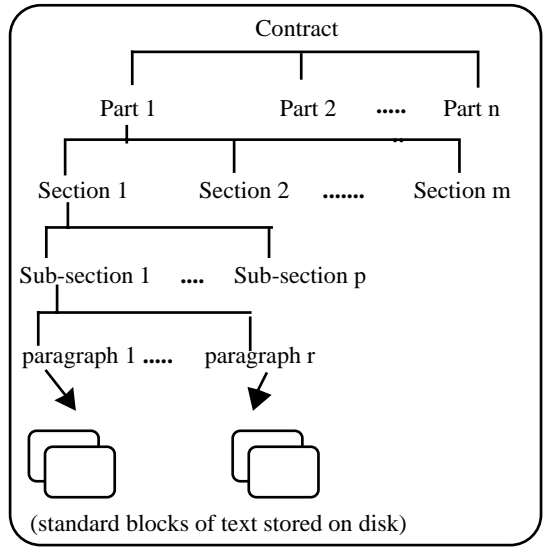

The actual document can be reconstructed in its entirety from this skeletal representation by a simple program that retrieves the appropriate fragments of text, instantiates any parameters with the specific data values, and collates the fragments into the final document. Currently the output document is plain text. We are exploring the possibility of producing documents in some mark-up language (specifically SGML [Bryan 1988], [Smith & Stutely 1988]) as our collaborators are interested in this.

A natural question for such representations is what document unit to take as the basic building block. For the sample contracts we have been dealing with, the section seems to be the most appropriate unit generally—but a feature of our system is that we do not need to commit to any particular choice of unit. In fact, different parts of the same document can be, and often are, represented at different levels of detail. In some cases it is unnecessary to subdivide the representation even to the level of individual sections; elsewhere, sections may be structured into sub-sections, sub-sections to paragraphs and sub-paragraphs, and sub-paragraphs to individual sentences. In principle, individual sentences could also be divided into phrases and other fragments but we have not bothered with this refinement since it would introduce text-processing requirements that would dominate the further development of the system.

Note also that the data values stored with document instances may be more than is specifically required to generate the actual document—for example it is often convenient to store the contracting parties' co-operation record, or other personal details even though they do not actually appear in the document itself. Similarly, document instances can be incorporated into a larger database recording details of the various parties and the projects and transactions in which they participate. We stress again that in this system we make no attempt to represent the content of any document unit in detail—other components not described here will cater for that.

Besides this skeletal representation, each document instance records other information. Keywords, which provide a simple mechanism by which users can search and retrieve documents, may be associated with any of the document's sub-units. (Here is an example of a simple technique which is practical for the retrieval of (certain types of) contracts but which is not usually adequate for indexing legislative texts in general.) Annotations by the drafter and general commentaries may also be associated with any sub-unit, and indeed the system is designed to encourage the drafter to record the reasons for any choices and modifications that are made.

Each document instance also contains a representation of textual cross-references between the units of text, and various other dependencies between them. Our industrial collaborators currently make use of hypertext systems for managing contractual documents and the cross-referencing information is intended to be used by the hypertext software when our system is eventually integrated. As explained later information about these dependencies is also used by the drafting system to check coherence of the document as it is created.

Annotations and some of the keywords need to be entered explicitly for each document instance, but much of this other information—commentaries and cross-references—is automatically inherited by all instances from what we call a *generic document*, a data structure representing what is common to all documents of some given type and recording all variations of previous document instances of the same type. In order to motivate the structure of a generic document, we need to say something about the nature of the drafting process we seek to support.

## 3  THE DRAFTING PROCESS

We want to support drafting of a contract based on existing examples of the same type. In engineering, it is standard practice for contracts to be drafted on the basis of model-form contracts, often issued by the relevant professional bodies. For example, many of the contracts that we have been examining are based on model-form contracts published by the Institution of Electrical Engineers (e.g. [IEE 1988, 1991]). Such model contracts have been developed over a considerable period—the first edition of [IEE 1988] was published in 1903. An important feature of these model-form contracts is that they are often accompanied by a detailed commentary, which explains the role of each individual provision in the document, its history and its overall effect. Where model-form contracts are available, they provide a natural starting point for a drafting system. However model-form contracts are not essential: where they are not available any previous document instance will do, although it is obviously most useful if some kind of commentary can be supplied for it.

To create a new document instance the user is provided with a model contract—a standard model-form contract or a previous contract of a similar type. Apart from changes in specific data values—or 'parameters'—many of the provisions will be acceptable in the original form. But there will also be sub-units or passages of the document which do not suit the circumstances at hand and which require some modification. In one example we have examined, Section 4-1 ('Precedence of Documents') of the model-form contract [IEE 1988] reads:

> Unless otherwise provided in the Contract the Conditions as amended by the Letter of Acceptance shall prevail over any other document forming part of the Contract and in the case of conflict between the General Conditions the Special Conditions shall prevail. Subject thereto the Specification shall prevail over any other document forming part of the Contract.

However, in the actual contract, a different text had been included in this section:

> The documents forming the Contract are to be taken as mutually explanatory of one another and in the case of ambiguities or discrepancies the same shall be explained and adjusted by the Engineer who shall thereupon issue to the Contractor appropriate instructions in writing.

In another example, Section 14-6 ('Rate of Progress') of the model-form contract [IEE 1988], which originally reads:

> The Engineer <u>shall</u> notify the Contractor if the Engineer <u>decides</u> that the rate of progress of the Works or of any Section is too slow to meet the Time for Completion and that this is not due to a circumstance for which the Contractor is entitled to an extension of time under Sub-Clause 33-1. [Emphasis added] had been modified to replace the occurrences of 'shall' and 'decides' by 'may' and 'considers' respectively. The point is that in neither case is there any indication as to why the modified version had been preferred over the original wording.

We want to provide a system in which the reasons for such modifications are recorded so that subsequent users can make informed choices about which version to select. We also want to allow users to create their own versions of sub-units where none of the existing ones is appropriate, and we want to encourage them to provide a commentary explaining the nature of the modification and the reasons for which it was made. Subsequent users will then be provided with several alternative versions—the original and the ones preferred in previously drafted documents—together with the accompanying commentary.

This collection of alternative versions is the core of the *generic document* for contracts of the given type. The model-form contract should not be confused with the generic document. A model-form contract may be used to construct the initial generic document, but thereafter the generic document grows as new document instances, containing new versions of sub-units, are created.

# 4 REPRESENTATION OF THE GENERIC DOCUMENT

We are now in a position to describe the representation of a generic document. Even if a contract does not contain explicit divisions into sub-agreements, we find it convenient to divide a contract into separate sub-agreements (or 'parts' for short), each of which collects together a number of related sections dealing with some aspect of the whole. Some of these 'parts' are *compulsory* in the sense that every document instance of this type must contain provisions dealing with this aspect of the agreement; other parts are *optional* in that the user can choose to include or omit them. The following list shows the 'parts' of the model-form contract [IEE 1988]. (There are also various appendices which we omit here for simplicity.) The optional parts are shown in square brackets:

> Definitions and Interpretations
>
> Engineer and Engineer's Representative
>
> [Assignment and Sub-Contracting]
>
> [Precedence of Documents]
>
> Basis of Tender and Contract Price
>
> [Changes in Costs]
>
> Purchaser's General Obligations
>
> Contractor's Obligations
>
> Suspension of Work, Delivery or Erection
>
> [Variations]
>
> [Defects Liability]
>
> Tests on Completion
>
> [Taking Over]
>
> [Performance Tests]
>
> Certificates and Payment
>
> [Accidents and Damage]
>
> Force Majeure
>
> [Insurance]
>
> [Disputes and Arbitration]
>
> Time for Completion

One might ask what determines whether a 'part' is compulsory or optional. Is it a legal requirement, or a professional standard, or perhaps common practice for one of the contracting parties? Is it simply an idiosyncrasy of an individual drafter? Our system does not make a distinction. All of these could be reasons for regarding a 'part' as compulsory. The treatment given to such 'parts' by the drafting system is the same independently of the reason, although the associated commentary may give some indication to the user of what this reason is.

Each 'part' is subdivided further, usually—but not necessarily—to a level corresponding to a 'section' of the document. And while sections have turned out to be the appropriate building blocks for the contracts we have been examining, sections can be further subdivided, to the required level of detail.

Any unit of the document (part, section, sub-section, etc.) can have several versions. At one time we allowed different versions of the same sub-unit to have different structures but we have since discovered that in practice it seems to be sufficient to use a much simpler scheme, where only the atomic units of text can have alternative versions. The following is an example of the representation of the generic document based on the model-form contract outlined earlier [IEE 1988]:

```
generic('IEE MF/2').

document_parameters('IEE MF/2',[$Engineer]).

part('IEE MF/2', 'Definitions and Interpretations',c).
part('IEE MF/2','Assignment and Sub-Contracting',o).
                    :
                    :
part('IEE MF/2', 'Time for Completion',c).

section('IEE MF/2', 'Time for Completion',
            'Extension of Time for Completion',      1,[]).
section('IEE MF/2', 'Time for Completion',
        'Delays by Sub-Contractors',
        2, []).
                    :
text_file('IEE MF/2',
     sect('Time for Completion',
          'Extension of Time for Completion'),
     1,[], tf1).

text_file('IEE MF/2',
     sect('Time for Completion',
          'Extension of Time for Completion'),
     2,[], tf2).
```

(This is standard Prolog syntax. The layout is simply to aid readability).

The predicate generic records that the type 'IEE MF/2' is a generic document type. The predicate document_parameters associates document types with parameters that are specific to them. In the example a value for $Engineer will be required when a specific document instance of this type is drafted. The actual IEE MF/2 requires a number of other parameters besides Engineer which we have omitted for

simplicity. Values for parties and a date for the agreement are common parameters for all documents and so they do not have to appear in the list of parameters specific to some document type. The part assertions record the 'parts' making up a given document type and whether they are compulsory ('c') or optional ('o'). The predicate section records for each 'part' of the given document type the sections contained within it. Identifiers for sections can be chosen arbitrarily but we tend to use the keywords, section titles or margin notes if they are present in the document. The fourth argument corresponds to the number of the section in the given part. This is not strictly necessary but it is convenient to have this when we generate the actual document text. The fifth argument is the list of parameters (here none) of the given section. Such parameters can be associated with any sub-unit of the document, at any level, but we have found that they are most useful when associated with the whole document and with the tips of the structure (which are usually sections in our example contracts). The predicate text_file associates atomic document units (here sections) with their respective text files. Here sect is a function symbol used to construct a document unit identifier. The third argument corresponds to the *version* of the given section, the fourth is its list of parameters and the last is a pointer to the file that contains the actual text.

This representation of the generic document is used by the drafting program to guide the creation of a document instance. An example of a document instance is as follows:

```
doc('IEE MF/2',
    'Q6',
    parties(...),
    date(...),
    [$Engineer = 'Frank'],
    [(sect('Definitions and Interpretations',
           'Singular and Plural'), 1),
     <other sections>])
```

The first argument shows the document type; the second is the identifier by which the document instance is referred to by the system (this is unique and automatically constructed); the third argument carries details about the parties and the fourth about the date of the contract; the fifth is the list of parameters and their values for the given document type. The sixth is a list of provisions, where each is a pair of a section identifier and the corresponding *version* used in the particular document instance. Hence, the document instance carries information about which *version* of a document sub-unit was chosen at the time of drafting.

The representation scheme provides some flexibility. Consider the following example [IEE 1988] of Section 38-1 ('Contractor's Equipment'):

> The Contractor shall within [30] days after the Letter of Acceptance provide to the Engineer a list of the Contractor's Equipment that the Contractor intends to use on the Site.

This can be represented in a number of different ways. Let §38-1 stand for the section identifier for readability. One view is that different values for "[30]" give rise to different versions of the section, in which case the representation would be:

text_file('IEE MF/2', §38-1, 1, [], tf1). text_file('IEE MF/2', §38-1, 2, [], tf2).

Another possibility is to treat "[30]" as a *parameter* value, and in this case, our representation would take the form:

text_file('IEE MF/2',§38-1,1,[$days=30],tf1).

If instead of the original section, the wording had been as follows:

> The Contractor shall provide to the Engineer a list of the Contractor's Equipment that the Contractor intends to use on the Site. Such list will be provided within [30] days after the Letter of Acceptance.

then we could represent this section in a third way, as two sub-sections, one for each sentence. The first sentence would have one version while the second would have (a) alternative textual versions or (b) a parameter value, as in the example above. Note however that this last option is difficult if the original wording is maintained. In this case, we would need more complicated text-processing mechanisms for interleaving sentence fragments.

As noted already, the drafter can create his own wording for (atomic) sub-units of the document, and our system provides a simple text editor for this purpose. However, it is the drafter's responsibility to ensure that the new text is meaningful and has the same properties as that which it replaces. The drafter is also allowed to extend the commentary and adjust the keywords corresponding to the section he modifies. However, if more dramatic modifications are required, which concern the structure of the document, then knowledge of the internal representation is demanded. Such modifications can be made but they are not supported by the drafting system at present.

The commentary, which is associated with the various sub-units, is recorded with the generic document and inherited by all instances. (Consequently, as a user creates new versions with additional commentary this becomes available automatically to all previous document instances as well. This seems a useful feature but we are considering a more complicated scheme in which the commentary provided for a document instance will be that which was available at the time of drafting.)

Keywords associated with each document instance can be adjusted by the user, and so, unlike commentary, they are recorded with the document *instance*. Keywords, which are stored with the generic document, are made available to the user during the drafting session as an initial suggestion. The system also provides a facility to record personal notes for each document instance.

# 5 CONSTRAINTS

We have said that document instances are stored with information about the cross-references (and other dependencies) between sub-units of the document. These are recorded as part of the generic document by means of assertions of the form:

$$\text{refers}(DocType, section_i, section_j).$$

The second and third arguments are identifiers for the units of text. These cross-references are inherited by all document instances.

Cross-reference information is compiled as part of the initial generic document representation and in the current system this must be done by the programmer manually. Wherever there is a textual (or other) reference from one sub-unit to another this fact must be recorded by the programmer using a refers assertion. Compiling this information is not as difficult as it might sound but it obviously requires going through the document in detail. We intend to develop utilities by which we can elicit such information (semi-) automatically. Cross-references are checked by the drafting module, as illustrated in the example drafting session in section 6.

The cross-references mentioned above are a kind of constraint that governs the structure of a document: If section **m** refers to section **n** and section **m** is included in the document instance, then section **n** should be included also. A feature of our system—to our eyes the most novel one—is that a wider range of explicit constraints on document structure is supported. We identify three kinds of constraints:

1. Constraints that link sub-units of the same document.

2. Constraints between data items/parameters.

3. Constraints relating data items/parameters and sub-units    of the document.

## 5.1    Constraints that link sub-units of the same document

In the generic document, we distinguish between optional and compulsory parts. It may be the case that two parts A and B are both optional, yet there may be a constraint that, if A is included in the document instance, then part B becomes compulsory. A specific example [IEE 1991] is where there is a part on Sub-Contracting and Assignment and a part on Sub-Contractors' Liability. Both are optional, but if the first is included then the second becomes compulsory.

Constraints of this general category, which relate document units to document units, are expressed using a simple special-purpose language. Assertions of the form:

forces(*DocType*, *A*, *B*).

are used to represent constraints of the form 'if document unit *A* is included then document unit *B* must also be included';

incompatible(*DocType*, *A*, *B*).

expresses that both *A* and *B* cannot appear in the document instance (alternatively 'if *A* is included then *B* must not be included');

exclusive_or(*DocType*, *A*, *B*).

indicates that exactly one of *A* and *B* must be included.

These three forms of constraints do not exhaust all the logical possibilities but we have not encountered examples where a more complex language would be necessary. (Notice that this simple language is quite expressive already since the specification of constraints may be subject to further conditions, expressed by Prolog clauses of the form:

forces(*DocType*, *A*, *B*)   :-   <*further conditions*>

and likewise for incompatible and exclusive_or.)

*5.2*   **Constraints between various data items**:

We also support constraints between data items. A simple example is a provision of the form 'if work is suspended for more than *three* months then payment is suspended for more than *six* months'. These constraints, which are specific to particular kinds of documents, are comparatively rare. Although we have made allowance for them in our system, we have not had to use them in practice.

More common are constraints between data items applying to contracts in general. The most obvious of these constraints is the requirement that contracting parties must not be identical.

**5.3**   **Constraints relating data/parameters and sub-units of the document**

The third type of constraint is used to deal with the case where values of various data items can affect the contents of what appears in the document. An example from [IEE 1991] is a requirement that if the party who supplies the service operates from outside the UK, then the document must include provisions stating arrangements for payment in foreign currency. Constraints of this type are expressed using the same language employed for constraints linking sub-units of a document. Thus, the given example would be represented according to the scheme:

forces(*DocType*, *Data*, *Document_Unit*).

For the example:

forces('IEE MF/2', [$Party-Address \= UK,

    part('Foreign Currency Payments')).

It should be noted that constraints of all three types are recorded with the generic document in such a way that they are automatically inherited by all the instances. Constraints of all kinds, including cross-references, are checked during the drafting process, either after every inclusion of a document unit or data item, or more usually, whenever the user explicitly requests it.

## 6    AN EXAMPLE OF A DRAFTING SESSION

This section provides a sample session of the drafting program and illustrates how the representations are used. The program is implemented in MacProlog: Prolog allows for efficient prototyping and is ideal for implementing the constraint-checking component. The MacProlog environment supplies a number of very useful primitives for the construction of a usable interface [LPA 1992].

Normally the drafting of a document is spread over several sessions but for this presentation, we will imagine that it is done in one session. During the drafting session, the user provides specific data values and makes choices about the contents of the document, which he drafts. The drafting module uses the input provided by the user and the generic information that we have stored in order to construct a document instance, while checking it against the constraints that are imposed on the whole process.

A danger in drafting large documents is that the user can become disoriented in the detail and large number of steps that need to be taken. In order to impose some structure on the process we have adopted a specific order in which drafting is performed. The user is permitted to go back and modify or adjust previous choices. Constraint checking can be activated or de-activated during the session from a menu. Thus, the user can choose whether checking takes place in a step-by-step fashion and/or at the end of the drafting session before the instance is actually stored.

A user begins the drafting session by making a selection from the database of available generic documents. A unique identifier for the new document instance is automatically constructed; the user can supply his own choice of a name by which the new document will be referred.

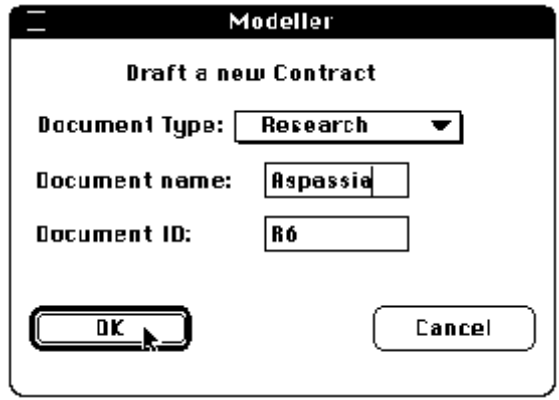

At some stage during the drafting (often but not necessarily at the beginning) the user is required to input information about the contracting parties (names, addresses) and the date the agreement is drafted.

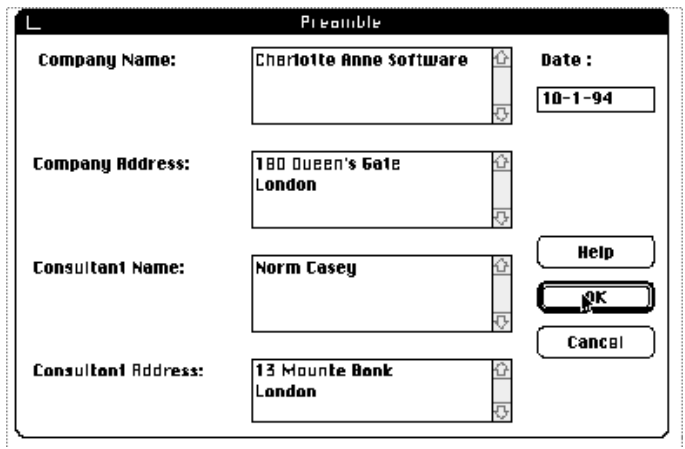

The associated help files simply explain what kind of information is required and where it typically appears in documents of this type. (Some of the IEE model contracts also provide some relevant commentary, which is made available to the user in the same way.)

Compulsory parts for the selected document type are presented in a menu. A user who is familiar with the contents of a part may choose to have it included automatically in the document, without going into it in detail.

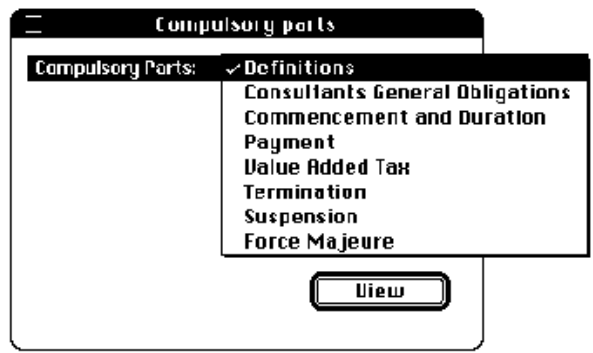

If the user prefers to view the contents of the selected part—or if a part chosen to be included automatically contains alternative versions or requires data values—the sections that it contains are presented one at a time, in a manner described below.

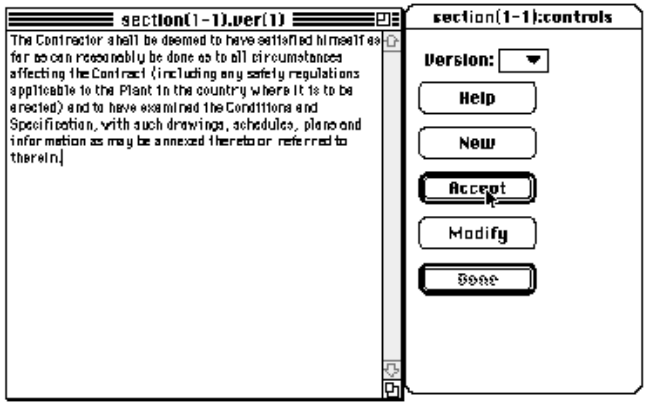

If there are alternative versions of the section then these can be displayed and compared. The *Help* button accesses the associated commentary and notes. If the user accepts an existing version of the section, a note is made in dynamic memory and the session continues with the next section or the next part. If no existing version is satisfactory, the user can create his own, possibly by modifying an existing one. A simple text editor is provided for this purpose. (As indicated in the previous sections this version of the system does not support more drastic modifications affecting the structure of the document, since this requires knowledge of the internal representation of documents. The user can however change the order in which sections or parts appear in the document.) The keywords associated with the section may also be modified.

Once compulsory parts have been dealt with, the user normally proceeds with optional parts in similar fashion, except that at this stage he has the opportunity to indicate that the contents of the document are now complete.

The user can check the document against the constraints at any point of the session or he can choose to have them checked automatically after every new entry.

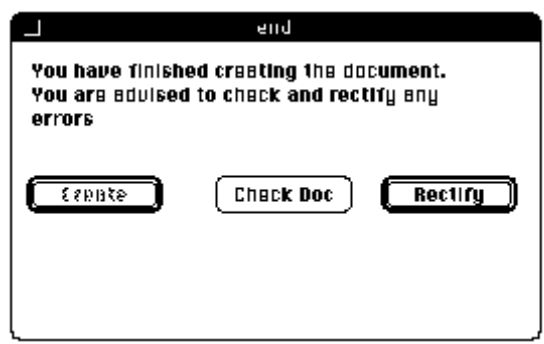

If violations occur, appropriate warning messages are displayed and remedial action is recommended.

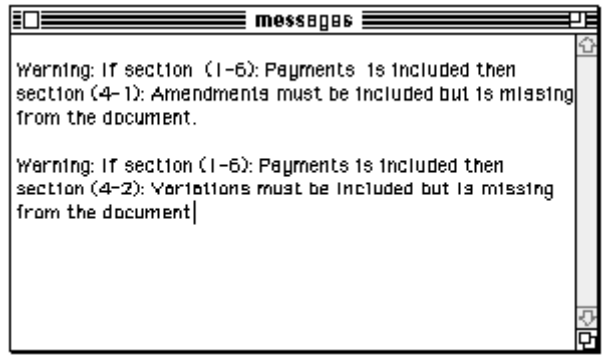

The prototype system also supports querying of the database of document instances.

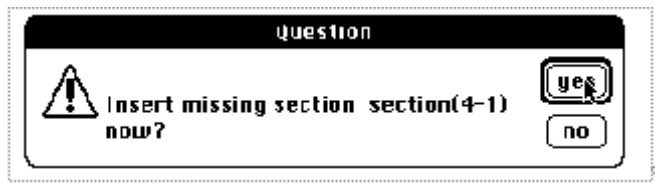

The less information the user provides the more general the query. Users can make specific queries, for example, stating the category of documents they are interested in, a date on, before or after which the documents were drafted, or they can provide particular information about the parties or the contents of the documents they are searching. An example of such a query is, "Find all contracts for research, which were drafted before December 1994, where the company has contracted with a party based in France, which contain version 3 of payment terms".

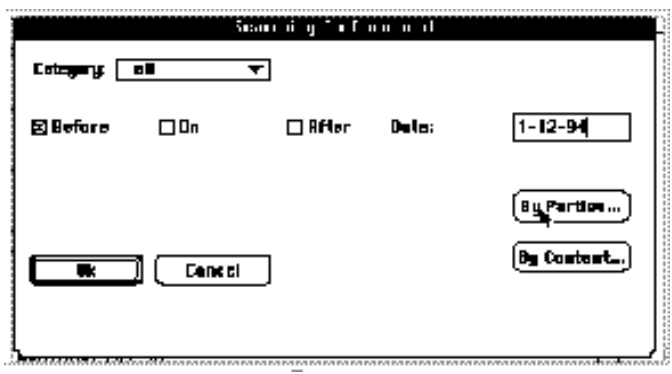

The user supplies additional information about parties and contents by selecting the appropriate buttons:

Documents that satisfy the requirements set out in the query are presented to the user as shown in the following picture and he can then select the ones he wishes to examine. Text files that correspond to the selected document(s) are accessed and manipulated by a separate module, allowing the actual text of the document to be viewed in its entirety.

## 7  CONCLUSIONS AND FUTURE WORK

We presented a framework within which the structure and coarse content of contracts can be represented for the purposes of drafting and contract database retrieval. The main features of our approach are:

1. The system addresses the structure and the interrelationships between the constituent parts of contracts, but not the text of the documents itself.

2. The representation of documents is separated from the mechanisms that manipulate it.

3. The drafting process is subject to a collection of explicitly stated constraints that govern the structure of the documents.

The prototype system is usable in its current form but it obviously needs further development and additional features—more sophisticated text editors, improved file management, interface with the hypertext systems, and so on—before it provides the facilities expected of a finished product. At this stage, our primary concern is to identify the main features to be supported. So far, we have tried the system on four contract types. (They are all from similar areas but we are confident that the approach is applicable to a wide class of contracts.)

We presume that one of the most common kinds of queries in which users might be interested is to determine what specific obligations and rights are imposed by the agreement on the contracting parties or what procedures apply under given circumstances. We have experimented with an extended representation, which indexes sub-units of a document not just by keywords, but also by recording explicitly 'duties' and 'rights' for each of the contracting parties. (This representation is just a form of indexing and is not an attempt to represent the meaning of terms such as 'duty' or 'right'.) This seems natural but it is still not clear whether users will find such a facility useful in practice.

As Lauritsen [1992] notes document assembly programs so far have been concerned mainly with automating the selection of document components, the insertion of specific values in selected textual templates and formatting work such as paragraph numbering and pagination. He stresses that "[c]hoosing what components to include in the first place, shaping them to extra-documentary objectives, and attending to semantic, strategic, and stylistic entailments among those choices, are where the real expertise—and the AI challenge—lies."

Of course, Lauritsen is right. Contract drafting involves much more than merely deciding which blocks of text to include in a new agreement. The system described in this paper is intended to provide the basic framework to which other tools dealing with other aspects of the drafting process will be attached.

However, the importance of contract assembly should not be underestimated either. It is not that we have chosen this approach as a temporary measure until we can develop something more sophisticated. In some circumstances contract drafting can best be viewed as an assembly process. What is important then is to make the dependencies and the effects of the various units as explicit as possible. We see the constraints described in this paper as a step in that direction.

**ACKNOWLEDGEMENTS**

Aspassia Daskalopulu is sponsored by a British Gas scholarship. We are grateful to Paul Cartledge, Philip Colby, Tony Fincham, Mark Green, Patrick Leonard and John Piggott of British Gas for providing us with the opportunity to work on this project and for their helpful advice, comments and assistance throughout its course.